\documentclass[10pt,twocolumn,letterpaper,table]{article}

\usepackage[dvipsnames, table]{xcolor}

\definecolor{lgreen}{RGB}{236, 255, 201}
\definecolor{nvgreen}{RGB}{118, 185, 0}
\definecolor{cvprblue}{rgb}{0.21,0.49,0.74}
\definecolor{lgray}{RGB}{245,245,245}

\definecolor{lightblue}{rgb}{0.678, 0.847, 0.902}
\definecolor{coral}{rgb}{1.0, 0.498, 0.314}
\definecolor{mint}{rgb}{0.686, 0.933, 0.651}
\definecolor{lavender}{rgb}{0.902, 0.902, 0.980}
\definecolor{peach}{rgb}{1.0, 0.855, 0.725}
\definecolor{teal}{rgb}{0.0, 0.502, 0.502}
\definecolor{gold}{rgb}{1.0, 0.843, 0.0}
\definecolor{skyblue}{rgb}{0.529, 0.808, 0.922}
\definecolor{salmon}{rgb}{0.980, 0.502, 0.447}
\definecolor{orchid}{rgb}{0.854, 0.439, 0.839}
\definecolor{plum}{rgb}{0.867, 0.627, 0.867}
\definecolor{chartreuse}{rgb}{0.498, 1.0, 0.0}
\definecolor{crimson}{rgb}{0.862, 0.078, 0.235}
\definecolor{khaki}{rgb}{0.941, 0.902, 0.549}

\usepackage{amsmath}
\usepackage{pifont}
\usepackage{float}
\usepackage{arydshln}
\usepackage{algorithm}
\usepackage{algorithmicx}
\usepackage{algpseudocode}
\usepackage{subcaption}
\usepackage{cuted}
\usepackage{bm}
\usepackage{dsfont}
\algrenewcommand\alglinenumber[1]{\tiny #1:}
\DeclareMathOperator*{\argmax}{arg\,max}

\usepackage{cvpr}              

\definecolor{cvprblue}{rgb}{0.21,0.49,0.74}

\usepackage{graphicx}
\usepackage{booktabs}
 
\newcommand\method{MDP}

\usepackage[pagebackref,breaklinks,colorlinks,allcolors=cvprblue]{hyperref}

\def\paperID{2901} 
\def\confName{CVPR}
\def\confYear{2025}

\title{MDP: Multidimensional Vision Model Pruning with Latency Constraint}

\author{Xinglong Sun, Barath Lakshmanan, Maying Shen, Shiyi Lan, Jingde Chen, Jose M. Alvarez\\
{NVIDIA}
}

\begin{document}
\maketitle

\begin{strip}
    \centering
    \begin{minipage}{.33\textwidth}
        \centering
        \text{Pruning \textbf{\textcolor{coral}{CNNs}} on \textbf{\textcolor{gold}{ImageNet}}}
        \includegraphics[width=\linewidth]{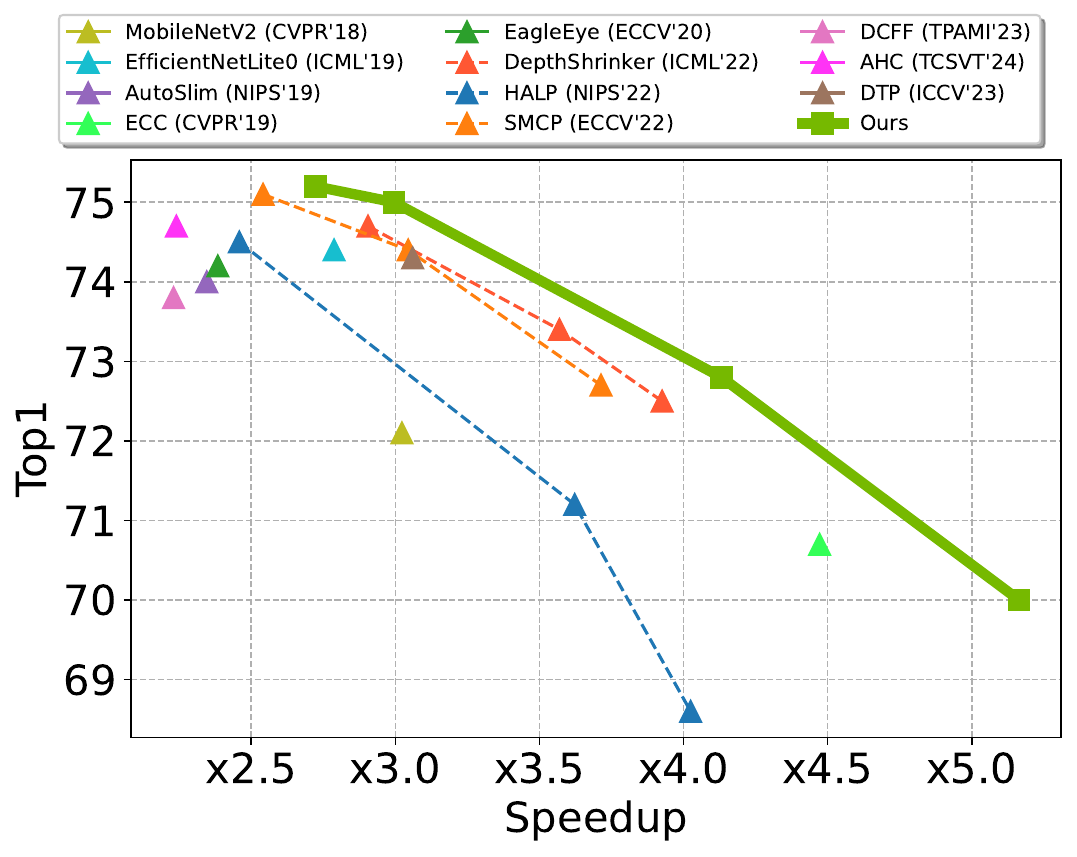}
    \end{minipage}
    \hfill
    \begin{minipage}{.33\textwidth}
        \centering
        \text{Pruning \textbf{\textcolor{teal}{Transformers}} on \textbf{\textcolor{gold}{ImageNet}}}
        \includegraphics[width=\linewidth]{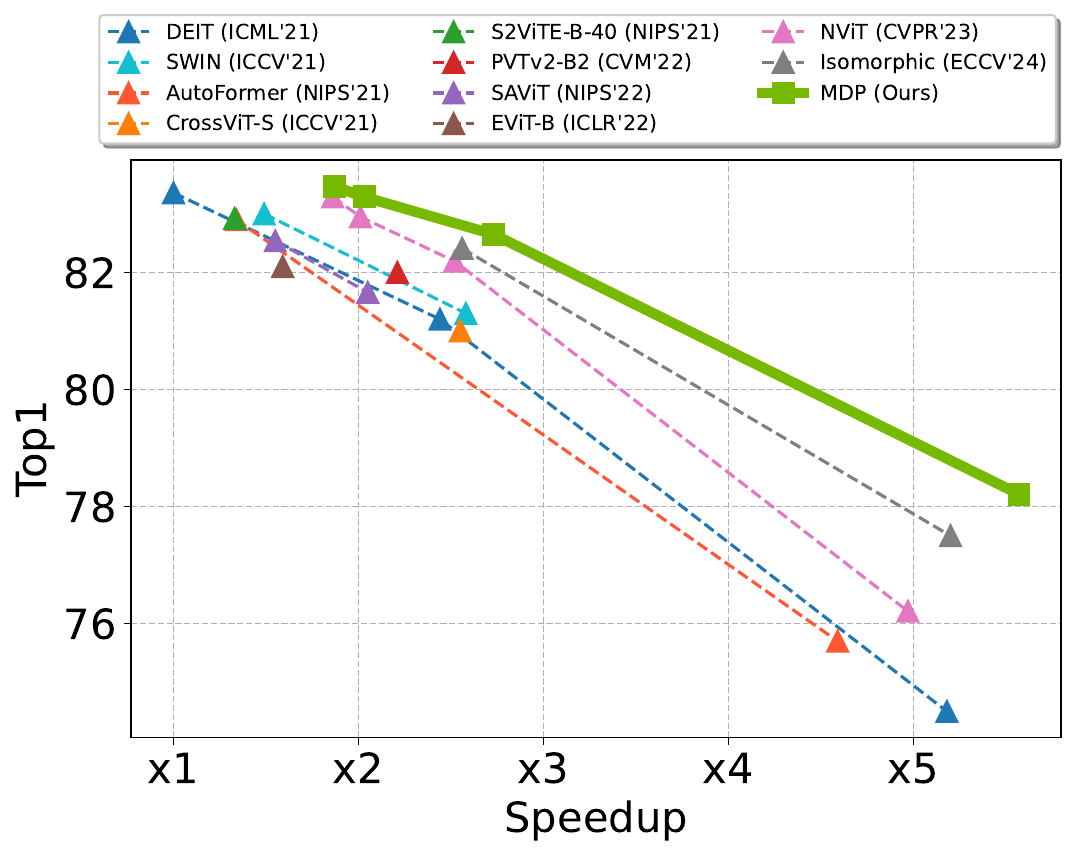}
    \end{minipage}
    \hfill
    \begin{minipage}{.33\textwidth}
        \centering
        \text{Pruning \textbf{\textcolor{plum}{3D Detection}} Model on \textbf{\textcolor{skyblue}{NuScenes}}}
        \includegraphics[width=\linewidth, height=130pt]{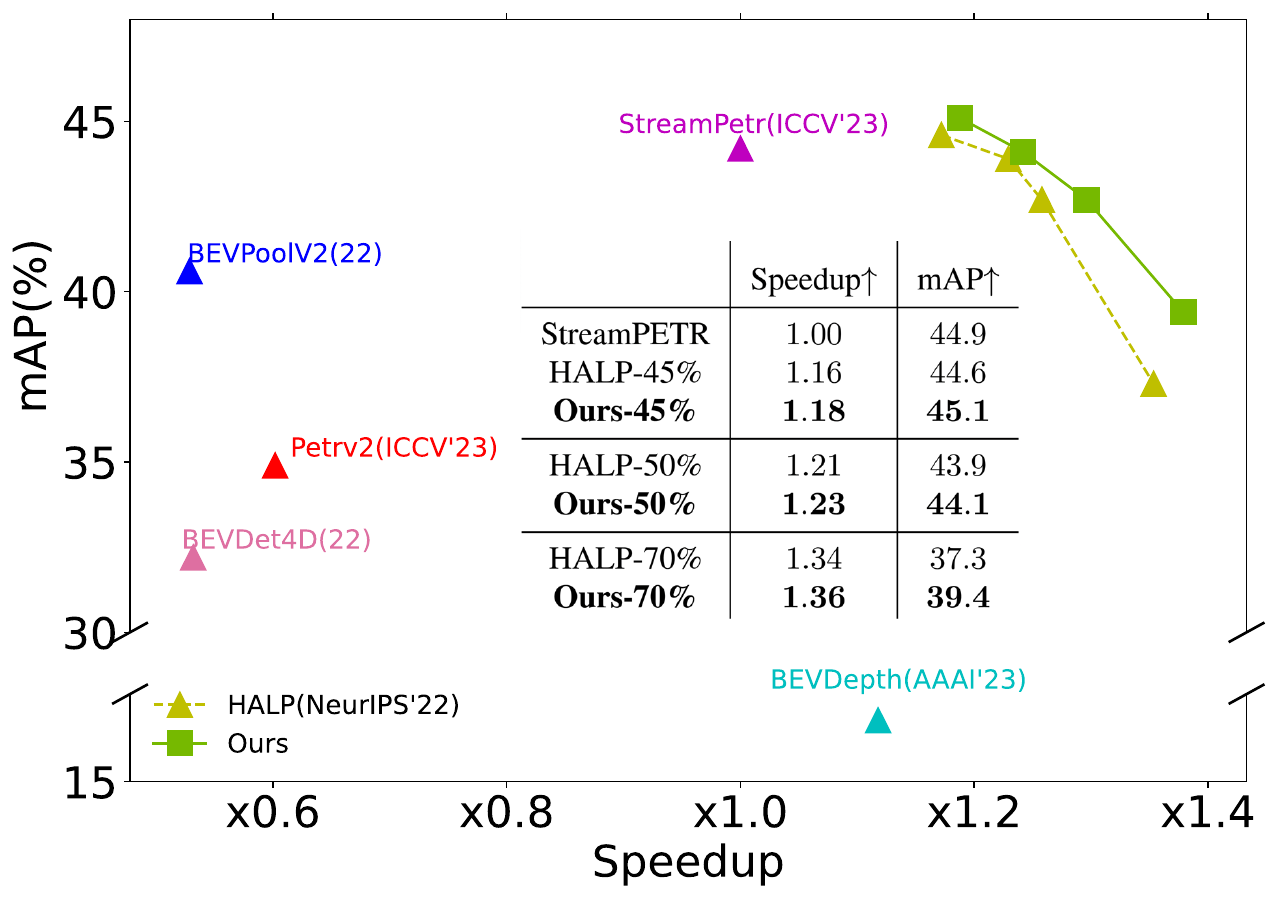}
    \end{minipage}
    \captionof{figure}{\method{} exhibits Pareto dominance with both CNNs and Transformers in tasks ranging from ImageNet classification to NuScenes 3D detection. \textit{Speedup are shown relative to the dense model.} \textbf{[Left]} On ImageNet pruning ResNet50, we achieve a $28\%$ speed increase alongside a $+1.4$ improvement in Top-1 compared with prior art~\cite{shen2021halp}. \textbf{[Middle]} On ImageNet pruning DEIT-Base, compared with very recent Isomorphic Pruning\cite{fang2025isomorphic}, our method further accelerates the baseline by an additional $37\%$ while yielding a $+0.7$ gain in Top-1. \textbf{[Right] }For 3D object detection, we obtain higher speed ($\times 1.18$) and mAP ($\mathbf{0.451}$ vs. $0.449$) compared to the dense baseline.} 
    \label{fig:teaser}
\end{strip}

\begin{abstract}
Current structural pruning methods face two significant limitations: (i) they often limit pruning to finer-grained levels like channels, making aggressive parameter reduction challenging, and (ii) they focus heavily on parameter and FLOP reduction, with existing latency-aware methods frequently relying on simplistic, suboptimal linear models that fail to generalize well to transformers, where multiple interacting dimensions impact latency. In this paper, we address both limitations by introducing Multi-Dimensional Pruning(MDP), a novel paradigm that jointly optimizes across a variety of pruning granularities—including channels, query/key, heads, embeddings, and blocks. MDP employs an advanced latency modeling technique to accurately capture latency variations across all prunable dimensions, achieving an optimal balance between latency and accuracy. By reformulating pruning as a Mixed-Integer Nonlinear Program (MINLP), MDP efficiently identifies the optimal pruned structure across all prunable dimensions while respecting latency constraints. This versatile framework supports both CNNs and transformers. Extensive experiments demonstrate that MDP significantly outperforms previous methods, especially at high pruning ratios. On ImageNet, MDP achieves a 28\% speed increase with a +1.4 Top-1 accuracy improvement over prior work like HALP for ResNet50 pruning. Against the latest transformer pruning method, Isomorphic, MDP delivers an additional 37\% acceleration with a +0.7 Top-1 accuracy improvement.
\end{abstract}
\section{Introduction}
\label{sec:intro}
Deep neural networks have become the de-facto standards of advanced computer vision applications~\cite{he2016deep,dosovitskiy2020image,liu2016ssd,wang2023exploring,long2015fully,sun2023revisiting,sun2024refining,harley2024tag}. Both \textit{convolutional neural networks (CNNs)}~\cite{he2016deep, liu2016ssd} and \textit{vision transformers}~\cite{dosovitskiy2020image, wang2023exploring} have demonstrated exceptional effectiveness across a wide range of tasks. As the performance advances, the models swell in size correspondingly, containing millions or even billions of parameters~\cite{kirillov2023segment}. This growth in model size presents challenges for deployment on resource-constrained edge devices, hinders real-time inference tasks such as autonomous driving.

Structural pruning~\cite{molchanov2019importance, shen2021halp, yang2023global, fang2025isomorphic,sun2025advancing}, which reduces redundant parameters in a structured manner—such as removing channels~\cite{molchanov2019importance}, attention heads~\cite{michel2019sixteen}, or specific components like queries and keys~\cite{yang2023global}—has proven to be an effective approach for decreasing model size and computation to meet real-time requirements without substantial accuracy loss. Despite its promise, structural pruning still faces two key limitations.

\textbf{Firstly (i)}, to keep up with the rapid increase in model sizes, we often need \textit{highly aggressive pruning techniques} to significantly reduce latency for efficient, real-time deployment. Most structural pruning methods currently target moderate pruning ratios (typically 30\%-50\%). These methods generally focus on removing channels or transformer components such as queries and heads, which alone are insufficient to reach the higher pruning ratios required (70\%-90\%). Achieving such high ratios often necessitates the structural removal of entire blocks or layers. While a few approaches~\cite{xu2020layer, wu2023block, tang2023sr, wang2019dbp, elkerdawy2020filter, chen2018shallowing} have explored layer and block pruning, they can only operate on the layer or block levels and cannot simultaneously integrate finer-grained sparsity, such as channel sparsity, leading to suboptimal accuracy.

\textbf{Secondly (ii)}, while many pruning methods focus on removing parameters~\cite{li2017pruning, molchanov2019importance, wang2021neural, lin2020hrank, fang2023depgraph} or reducing FLOPs~\cite{li2020eagleeye, wu2020constraint, yu2019autoslim}, recent studies~\cite{tan2019mnasnet, shen2021halp} have shown that both these metrics don't directly correlate with inference latency, challenging practical speedups. Some recent approaches~\cite{shen2021halp, humble2022soft,  shen2023hardware} have begun exploring hardware-aware pruning that adheres strictly to an inference latency budget. However, they rely on a suboptimal latency model that assumes a linear relationship with channel counts—effective only in CNNs, often imprecise, and unsuitable for transformers where interacting multiple prunable dimensions make this linear model completely infeasible.

In this paper, we propose a paradigm that simultaneously addresses challenges \textbf{(i)} and \textbf{(ii)}. We identify the key limitation of prior methods as their inability to handle multiple pruning dimensions simultaneously. This multidimensionality naturally arises when combining various granularities—such as block, layer, and channel sparsity—or when accurately modeling latency, which is impacted by multiple dimensions, such as embedding, heads, Q/K/V, and MLP sizes in transformers. Specifically, for any vision model, we first identify and encode the prunable dimensions with variables. To enable block-level pruning, we model block removal decisions as an additional dimension, grouping parameters within the same block for joint handling in the optimization process. For accurate latency modeling to guide pruning, we construct a latency constraint function that considers all prunable dimensions affecting latency, thereby capturing the simultaneous variations across them. Together, we innovatively reformulate pruning as a Mixed-Integer Nonlinear Program (MINLP) \cite{lee2011mixed, burer2012non, bussieck2003mixed} optimization problem, allowing us to directly solve for the globally optimal pruned structure across all dimensions while strictly adhering to a latency budget. All together, we refer to our method as \textbf{M}ulti-\textbf{D}imensional \textbf{P}runing (\method{}). \textit{Our code will be provided upon acceptance.}

\begin{figure*}[t!]
\vspace{-5pt}
    \centering
    \includegraphics[width=\linewidth]{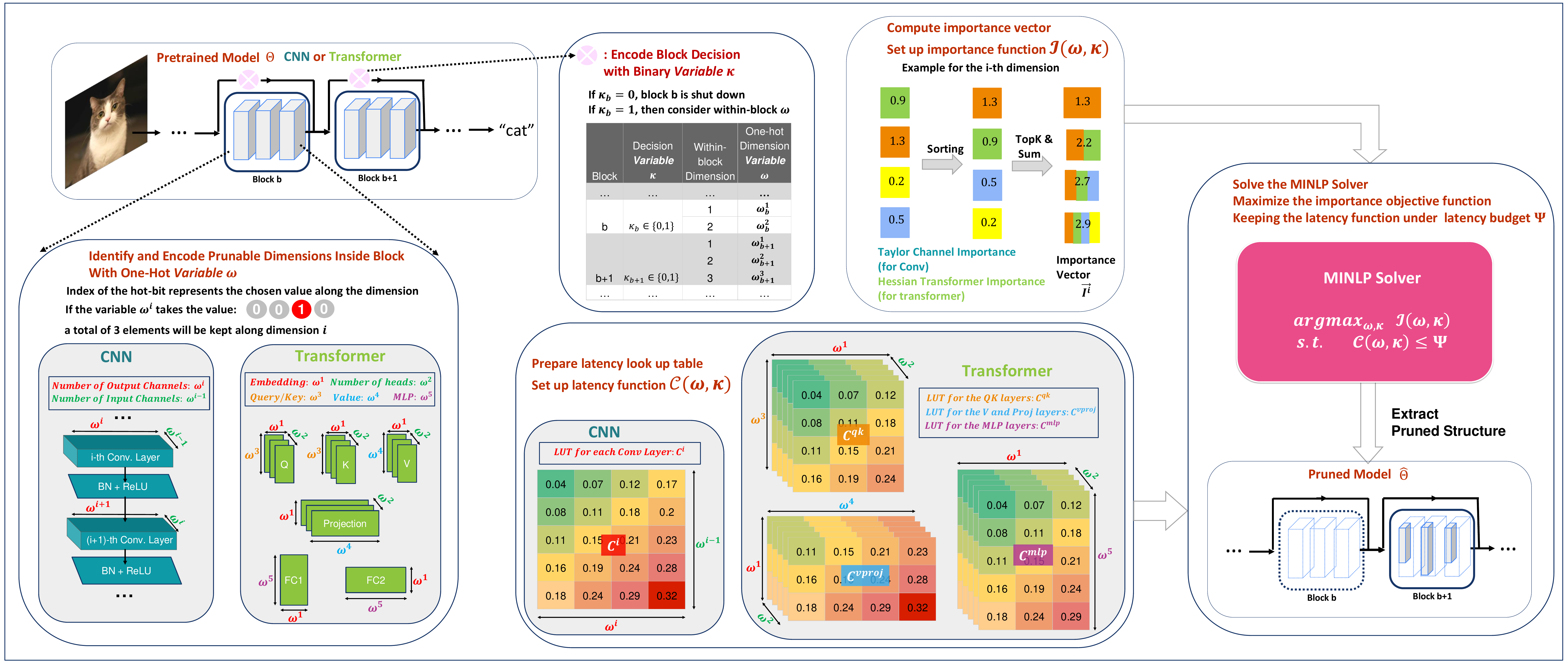}
    \caption{We begin by encoding prunable dimensions within the model with one-hot variables($\bm{\omega}$), followed by establishing an importance objective and a latency constraint for each value of $\bm{\omega}$ with prepared latency lookup table (LUT). Next, parameters are grouped by block, and an MINLP optimizes pruning across all dimensions under latency budget $\Psi$. Finally, we extract the pruned subnetwork and finetune it.}
    \label{fig:paradigm}
    \vspace{-10pt}
\end{figure*}

Our extensive experiments, with a glimpse offered in Figure~\ref{fig:teaser}, demonstrate the state-of-the-art performance of our method across diverse scenarios, including pruning CNNs and transformers for tasks ranging from classification to 3D detection. Our method’s advantage is especially clear \textit{at high pruning ratios}. For example, when pruning CNNs with an aggressive 85\% ratio, we surpass the previous latency-pruning art HALP~\cite{shen2021halp}, achieving a $28\%$ speed increase alongside a $+1.4$ improvement in Top-1 accuracy. Compared to the recent transformer pruning work Isomorphic~\cite{fang2025isomorphic}, our method further accelerates the dense DEIT-B baseline by an additional $37\%$ while yielding a $+0.7$ gain in accuracy. For 3D object detection, we set a new benchmark by pruning StreamPETR\cite{wang2023exploring} by $45\%$, achieving superior speed ($\times 1.18$) and mAP ($\mathbf{0.451}$ vs. $0.449$) compared to the dense baseline.

We summarize our contributions as follows:
\begin{itemize}
    \item We enable pruning with multiple structural granularities, such as simultaneous channel and block pruning, allowing collective decision-making in optimization.
    \item We propose an accurate latency formulation to guide hardware-aware pruning, effectively capturing the impact of all prunable dimensions within the model.
    \item We integrate these strategies with a novel paradigm, redefining pruning as a Mixed-Integer Nonlinear Program (MINLP) to solve for a globally optimal pruned structure across all dimensions within latency constraints.
    \item Our framework is versatile and applies to both CNNs and transformers, achieving the \textit{first latency-constrained transformer pruning} to our knowledge in the field.
\end{itemize}

\section{Related Works}
\label{sec:related}
Our work can be categorized as a pruning method in general. We will now provide a brief overview of the field and highlight our differences from the previous approaches. Pruning methods~\cite{lecun1990optimal, hassibi1992second, han2015deep, molchanov2017variational, sun2022disparse, alvarez2016learning, li2017pruning, shen2021halp, molchanov2019importance, lin2020hrank, sun2025advancing} mostly design importance criterion to rank parameters and remove the lowest-ranked ones.

\noindent\textbf{Channel Pruning}
Some pruning methods~\cite{li2017pruning, chin2020towards, he2020learning, he2018soft, yang2018netadapt, he2019filter, lin2020hrank, molchanov2019importance, sun2024towards, shen2021halp, humble2022soft, fang2025isomorphic, sun2023pruning} operate under structural constraints, for example removing convolutional channels\cite{li2017pruning} from CNNs, thus enjoy immediate performance improvement without specialized hardware or library support. Exemplary channel importance criterion relied on metrics like weight norm~\cite{li2017pruning, chin2020towards, he2020learning, he2018soft, yang2018netadapt}, Taylor expansion~\cite{lin2018accelerating, molchanov2019importance, you2019gate}, geometric median~\cite{he2019filter}, and feature maps rank~\cite{lin2020hrank}. Our method leverages the Taylor~\cite{molchanov2019importance} channel importance criterion but extend it to evaluate the configurations of entire layers and blocks, going beyond just pruning channel but also combining layer and block removals. 

\noindent\textbf{Transformer Pruning}
In recent years, numerous studies \cite{chen2021chasing, yang2023global, yu2022unified, kong2022spvit, zheng2022savit, zhu2021vision, fang2025isomorphic} have adapted structural pruning techniques for modern vision transformers \cite{dosovitskiy2020image, touvron2021training, chen2021crossvit, liu2021swin}. Transformer pruning has been explored through various approaches, including token pruning \cite{kong2022spvit}, width reduction \cite{zhu2021vision}, attention head removal \cite{michel2019sixteen}, and architecture transformation \cite{he2024pruning}. Recently, NViT \cite{yang2023global} introduced a Hessian-based importance scoring to enable global pruning of transformer components. Our method also uses Hessian-based scoring for individual transformer parameters. However, unlike prior methods focused on maximal parameter or FLOPs reduction, our approach prunes strictly under a specified inference latency budget, ensuring practical speedup.

\noindent\textbf{Layer and Block Pruning}
Most prior pruning methods are effective at minimizing performance loss with moderate pruning but struggle with more extensive pruning, as they focus on smaller structures (e.g., channels or queries/keys) rather than larger ones like layers or blocks. Only a few works \cite{chen2018shallowing, elkerdawy2020filter, wang2019dbp, tang2023sr, wu2023block, xu2020layer} explore pruning at the layer and block levels, though remaining limited and lack support for mixed-granularity pruning (e.g., combined with channel sparsity). Some approaches also add complexity with extra modules like linear probes \cite{wang2019dbp, chen2015compressing}. Our method unifies pruning across granularities, achieving optimal pruning at all levels without additional modules or training.

\noindent\textbf{Hardware-aware Pruning} 
Since parameter reduction does not directly translate to computational savings, some methods \cite{li2020eagleeye, wu2020constraint, yang2018netadapt} focus primarily on reducing model FLOPs. More recent approaches take this further with hardware-aware pruning, targeting a direct reduction in hardware inference latency. A key work, HALP \cite{shen2021halp}, frames pruning as a knapsack problem \cite{sinha1979multiple}, maximizing parameter importance while constraining latency within a specified budget. To improve the learning capacity of pruned models, SMCP \cite{humble2022soft} introduces soft masking into HALP’s framework, allowing pruned weights to be reconsidered iteratively.

Despite notable progress in accuracy and speed, these methods \cite{shen2021halp, shen2023hardware, humble2022soft} rely on an oversimplified latency estimation, resulting in suboptimal accuracy-latency trade-offs. They model latency linearly with respect to output channels in CNNs but overlook simultaneous input channel variations. Additionally, they do not extend to transformers, where latency is affected by multiple dimensions, including Q, K, V, head, embedding, and MLP sizes. While some hardware-aware pruning works exist for transformers \cite{kong2022spvit, yang2023global, molchanov2022lana}, including NViT \cite{yang2023global}, these approaches only use latency as a soft regularization term and cannot guarantee latency constraints. In our work, we also focus on hardware-aware pruning but introduce a more accurate latency modeling approach that captures simultaneous variations across \textbf{\textit{all dimensions}} affecting latency, enabling precise pruning to meet strict hardware constraints.

\noindent\textbf{Mixed-Integer Nonlinear Program (MINLP)~\cite{lee2011mixed, burer2012non, bussieck2003mixed}} 
Since our strategies are unified through a MINLP formulation, we briefly introduce this field. Formally defined in \cite{lee2011mixed}, MINLPs are optimization problems involving both integer and continuous variables, with nonlinear objective functions and constraints. Solving MINLPs efficiently \cite{bonami2008algorithmic, d2013mixed, gunluk2010perspective, duran1986outer, fletcher1994solving, bonami2009feasibility, bernal2020improving} is a key research area, often requiring decomposition into Mixed-Integer Linear Programs (MILPs) and Nonlinear Programs (NLPs). Recently, Python packages such as Pyomo\cite{bynum2021pyomo} and MindtPy~\cite{bernal2018mixed} have simplified modeling and solving MINLPs. In this work, we employ the Outer Approximation (OA) method~\cite{duran1986outer,fletcher1994solving} with Pyomo and MindtPy to solve our pruning MINLP.
\section{Methodology}
We will now present our pruning framework. We begin by
establishing preliminaries, defining our goals and declaring relevant notations. We then describe our proposed pruning formulation in detail. Since our framework is applicable to both convolutional and transformer models, we use generic notations and \textit{cover both cases throughout.}

\noindent\textbf{Preliminaries} 
Consider a pretrained model $\Theta$, our goal here is to find the \textit{most performant} subnetwork $\hat{\Theta} \subseteq \Theta$ through removing parameters, such that the inference latency of $\hat{\Theta}$ is below the desired latency budget $\Psi$ on the target hardware. Moreover, we define a \textit{block} as a structural group of parameters. Specifically, \textit{in CNNs}, following ~\cite{he2016deep, luo2020neural}, a block is defined as the set of layers(e.g. bottleneck~\cite{he2016deep}) skipped by a residual connection; while \textit{in transformers}, following ~\cite{dosovitskiy2020image}, a block comprises the standard qkv self-attention, layer normalization, and two linear MLP layers, which are also typically bypassed by a residual connection. Suppose there is a total of $\mathcal{B}$ blocks in $\Theta$. Specifically \textit{for CNNs}, let $L_b$ denote the total number of convolutional layers within the $b$-th block.
\subsection{Multi-Dimensional Pruning (\method{})}
\label{subsec:ours}
Given a pretrained model, we first identify prunable dimensions and encode them with one-hot variables $\bm{\omega}$. We then construct an importance function, $\mathcal{I}(\bm{\omega})$, to assign importance scores to each possible value of $\bm{\omega}$. To guide pruning within a latency budget $\Psi$, we define a latency constraint function, $\mathcal{C}(\bm{\omega})$, that maps $\bm{\omega}$ values to their inference latencies. For efficient evaluation, we introduce a decomposed version, $\mathcal{C}'(\bm{\omega})$, which performs model-specific decomposition on $\mathcal{C}(\bm{\omega})$. Additionally, to integrate sparsity at the layer and block levels, we introduce a binary decision variable $\bm{\kappa}$, grouping all parameters within each block for joint optimization. We then formulate all the above into a Mixed-Integer Nonlinear Program (MINLP) to to maximize $\mathcal{I}(\bm{\omega}, \bm{\kappa})$ while constraining $\mathcal{C}'(\bm{\omega}, \bm{\kappa})$ under the latency budget $\Psi$, achieving an optimal structure across all dimensions. This method, termed \textbf{M}ulti-\textbf{D}imensional \textbf{P}runing (\textbf{MDP}), is illustrated in Figure~\ref{fig:paradigm}, with each step detailed below.

\noindent \textbf{Identify and Encode Prunable Dimensions} 
To begin, we need to determine which parameters in the model can be pruned and which cannot. Suppose there are a total of $\Omega$ independent prunable dimensions existing in the model, we represent all of them collectively as $\bm{\omega}$. For the $b$-th block, assume there are a total of $\Omega_b$ independent prunable dimensions, with the $i$-th dimension denoted by a \textbf{\textit{one-hot variable}} $\omega_b^i$ for $i \in [1, \Omega^b]$, where the index of the "hot" bit indicates the chosen value along that dimension by pruning.

Concretely for the $b$-th block:

\textbf{\textit{For Convolutions:}} Following established approaches of channel pruning~\cite{molchanov2019importance, shen2021halp, humble2022soft, lin2020hrank, sun2024towards}, the prunable dimension corresponds to the number of output convolution channels($c_{out}$) in each layer. We would have $\Omega_b=L_b$. For the $i$-th convolutional layer, its output channel count would be encoded by $\omega_b^i$ whereas the input channel count by $\omega_b^{i-1}$. For example if the one-hot bit index of $\omega_b^i$ takes $j$, we would have a total of $j$ kept output channels at layer $i$ after pruning.

\textbf{\textit{For Transformers:}} Drawing on recent transformer pruning methodologies~\cite{yang2023global}, we focus on five independent prunable dimensions per block: the embedding dimension ($emb$) $\omega_b^1$, the number of attention heads ($head$) $\omega_b^2$, the query/key dimension ($qk$) $\omega_b^3$, the value dimension ($v$) $\omega_b^4$, and the MLP hidden dimension ($mlp$) $\omega_b^5$. Here, we would have $\Omega_b=5$. For example if the one-hot bit index of $\omega_b^3$ takes $j$, we would have the query/key dimension reduced to $j$.

Together, the variables $\bm{\omega}$ define the structural configuration of all possible subnetworks $\hat{\Theta}$ that can be derived from the original model through pruning. The number of all possible $\hat{\Theta}$ is equal to $\prod_{b}^B\prod_{i}^{\Omega_b} |w_b^i|$. For further clarity, please refer to the graphical illustrations in Fig.\ref{fig:paradigm}.

\noindent \textbf{Construct Importance Objective Function} As discussed in Sec.\ref{sec:related}, the quality of the pruned subnetworks could be effectively assessed with importance scores. Our goal here is to construct an importance \textit{objective function} $\mathcal{I}(\cdot)$ which associate importance scores along each dimension with the dimension variables $\bm{\omega}$ defined above, allowing us to understand how selecting different values along each dimension influences model performance. Specifically, for each dimension variable $\omega_b^i$, we construct an \textit{importance vector} $\Vec{I_b^i}$ of equal length, i.e. $|\omega_b^i| = |\Vec{I_b^i}|$, and the $j$-th value of $\Vec{I_b^i}$ represents the importance of retaining $j$ elements along dimension $i$. Notably, our method is \textit{agnostic} to the choice of importance score, and here we use well-established scores specific to each model family as follows:

\textbf{\textit{For Convolutions:}} we first adopt the widely-known Taylor importance~\cite{molchanov2019importance} to score each channel; next, to construct $\Vec{I_b^i}$, for the $j$-th element of $\Vec{I_b^i}$, we aggregate the \textbf{\textit{Top}}-$j$ Taylor channel scores.

\textbf{\textit{For Transformers:}} we adopt the recent Hessian-aware importance~\cite{yang2023global} which evaluate parameter saliency globally across all dimensions; next, to construct $\Vec{I_b^i}$, we perform a similar aggregation, with each $j$-th element calculated by summing the \textbf{\textit{Top}}-$j$ Hessian scores.

Unlike traditional pruning methods, our approach does not assign scores to an \textit{individual channel or transformer parameter like query/key;} instead, each transformed importance vector $\Vec{I_b^i}$ represents the importance of keeping a given \textit{total number of channels or queries/keys}. Bringing all components together, we define $\mathcal{I}(\bm{\omega})$ as:
{\small
\begin{align}
    \label{eqn:importance}
    \mathcal{I}(\bm{\omega}) = \sum_{b}^B\mathcal{I}(\bm{\omega_b}) = \sum_{b}^B \sum_{i}^{\Omega_b} \omega_b^{i\top} \cdot \Vec{I_b^i}
\end{align}
}
\noindent \textbf{Construct Latency Constraint Function} 
To meet the desired latency budget, we need a latency constraint function, $\mathcal{C}(\cdot)$, that maps dimension vectors $\bm{\omega}$ to inference latency. We create a latency lookup table (LUT), precomputed on target hardware, allowing us to estimate inference latency given different values of $\bm{\omega}$. For each model block $b$, the corresponding table $\mathbf{C}_b$ provides latency estimates and is structured as $\mathbf{C}_b \in \mathbb{R}^{|\omega_b^1|\times|\omega_b^2|\times\hdots|\omega_b^{\Omega_b}|}$. The constraint function can thus be formulated as:
{\small
\begin{align}
    \label{eqn:before_decompose}
    \mathcal{C}(\bm{\omega}) &= \sum_b^B\mathcal{C}(\bm{\omega_b}) \nonumber\\
    &= \sum_b^B \textbf{\textsc{Sum}}\Big((\omega_b^1 \otimes \omega_b^2 \otimes \hdots \otimes \omega_b^{\Omega_b}) \odot \mathbf{C}_b\Big),
\end{align}
}
where $\otimes$, $\odot$ represent outer product, element-wise product, and \textbf{\textsc{Sum}} denotes computing the sum of all entries of the tensor. However, computing this directly is impractical due to the chain of sequential outer products and gigantic $\mathbf{C}_b$. To enable efficient computation, we perform model-specific \textbf{\textit{decomposition}} of Eqn.\ref{eqn:before_decompose} and the lookup tables $\mathbf{C}_b$. We denote $\mathcal{C}(\omega)$ after decomposition as $\mathcal{C}'(\omega)$ with detail shown below:

\textbf{\textit{For Convolutions:}} each block’s latency is decomposed as the sum of individual convolutional layers, defined by input and output channel count($\omega_b^{i-1}$ and $\omega_b^i$). The table $\mathbf{C}_b$ is decomposed into smaller ones $\mathbf{C}_b^i \in \mathbb{R}^{|\omega_b^{i-1}| \times |\omega_b^i|}$, giving:
{\small
\begin{align}
    \label{eqn:conv_latency}
    \mathcal{C}'(\bm{\omega}) &= \sum_b^{B}\mathcal{C}'(\bm{\omega_b}) \\
    &= \sum_b^B \sum_i^{\Omega_b} \textbf{\textsc{Sum}}\Big(\omega_b^{i-1} \otimes \omega_b^{i}) \odot \mathbf{C}_b^i\Big) \nonumber
\end{align}
}
\textbf{\textit{For Transformers:}} as depicted in Fig.\ref{fig:paradigm}, transformer block latency is decomposed into three parts: (i) QK layers, defined by \textit{head}, \textit{embedding}, and \textit{qk} dimensions ($\omega_b^1, \omega_b^2, \omega_b^3$) with table $\mathbf{C}_b^{qk}$; (ii) V and projection layers, defined by \textit{head}, \textit{embedding}, and \textit{v} dimensions ($\omega_b^1, \omega_b^2, \omega_b^4$) with table $\mathbf{C}_b^{vproj}$; and (iii) MLP layers, defined by \textit{embedding} and \textit{mlp} dimensions ($\omega_b^2, \omega_b^5$) with table $\mathbf{C}_b^{mlp}$. This gives:
{\small
\begin{align}
    \label{eqn:transformer_latency}
    \mathcal{C}'(\bm{\omega}) &= \sum_{b=1}^{B} \mathcal{C}'(\bm{\omega_b}) \\
    &= \sum_{b=1}^B \Bigg[\textbf{\textsc{Sum}}\Big((\omega_b^1 \otimes \omega_b^2 \otimes \omega_b^3) \odot \mathbf{C}_b^{qk}\Big) \nonumber\\
    &\quad + \textbf{\textsc{Sum}}\Big((\omega_b^1 \otimes \omega_b^2 \otimes \omega_b^4) \odot \mathbf{C}_b^{vproj}\Big) \nonumber\\
    &\quad + \textbf{\textsc{Sum}}\Big((\omega_b^1 \otimes \omega_b^5) \odot \mathbf{C}_b^{mlp}\Big) \Bigg] \nonumber
\end{align}
}
This decomposition enables efficient evaluation of $\mathcal{C}'(\bm{\omega})$ using modern numerical optimization tools. 

While previous latency pruning methods~\cite{shen2021halp, humble2022soft, shen2023hardware} also use latency lookup tables, they rely on \textbf{\textit{linear}} models that only account for output channel counts (e.g., capturing $\omega^i$ in convolutional layer $i$). Such formulations are not only imprecise(ignoring input channel dimension $\omega^{i-1}$) but also infeasible for transformers, where $\Omega=5$ prunable dimensions exist. In contrast, our latency modeling accurately incorporates changes across all prunable dimensions, providing precise and optimal latency guidance. More detailed analysis in latency modeling can be found in the appendix.

\noindent \textbf{Enable Block Removal} Structural pruning at higher granularities, such as layers and blocks, poses unique challenges. This complexity arises because arbitrarily pruning a single layer could easily lead to network disconnection, causing discontinuity in the information flow of the network. However, blocks are inherently resilient to removal of all their internal parameters at once, as the skip connection allows information to bypass the removed block, preserving gradient flow. 

To manage the removal of an entire block, we employ \textit{block grouping}, where the importance and latency constraints associated with each block are combined. If pruning removes the $b$-th block, then both the importance and latency contributions of all parameters within that block are set to zero. This group decision is modeled with binary block decision variables $\kappa_b$, which adjust the importance objective function $\mathcal{I}(\bm{\omega})$ and the latency constraint $\mathcal{C}'(\bm{\omega})$ to account for block removal as follows:
{\small
\begin{align}
    \label{eqn:block_importance}
    \mathcal{I}(\bm{\omega}, \bm{\kappa}) &= \sum_{b}^B\kappa_b \cdot\mathcal{I}(\bm{\omega_b})\\
    \label{eqn:block_latency}
    \mathcal{C}'(\bm{\omega}, \bm{\kappa}) &= \sum_b^B\kappa_b\cdot\mathcal{C}'(\bm{\omega_b})
\end{align}
}
\noindent \textbf{Solve MINLP} With $\mathcal{I}(\bm{\omega}, \bm{\kappa})$ and $\mathcal{C}'(\bm{\omega}, \bm{\kappa})$ defined, our goal is to find the optimal $\bm{\omega}$ and $\bm{\kappa}$ such that the importance objective function $\mathcal{I}(\bm{\omega}, \bm{\kappa})$ is maximized while the latency constraint function $\mathcal{C}'(\bm{\omega}, \bm{\kappa})$ stays below the latency budget $\Psi$. Concretely, we formulate \method{} pruning as a \textit{Mixed-Integer Nonlinear Programming (MINLP)} formulation:
{\small
\begin{align}
    \label{eqn:program}
\argmax_{\bm{\omega}, \bm{\kappa}} \quad \mathcal{I}(\bm{\omega}, \bm{\kappa}), \quad
\textsc{s.t.} \quad \mathcal{C}'(\bm{\omega}, \bm{\kappa})  \leq \Psi 
\end{align}
}
where each $\omega_b^i$ is a one-hot vector, $\kappa_b$ is binary, and both the importance vector $\Vec{I}$ (for the importance objective \ref{eqn:importance}) and latency lookup tables $\mathbf{C}$ (for the latency constraint \ref{eqn:conv_latency}, \ref{eqn:transformer_latency}) contain real-valued entries, resulting in a \textit{mixed-integer} program. Additionally, while $\mathcal{I}(\bm{\omega})$ Eqn.\ref{eqn:importance} is linear in $\bm{\omega}$, introducing the product with block decision variables $\bm{\kappa}$ makes $\mathcal{I}(\bm{\omega}, \bm{\kappa})$ Eqn.\ref{eqn:block_importance} quadratic. The latency constraint $\mathcal{C}'(\bm{\omega}, \bm{\kappa})$ varies in complexity, becoming cubic for \textit{CNNs}(Eqn.\ref{eqn:conv_latency}, \ref{eqn:block_latency}) and quartic for \textit{transformers}(Eqn.\ref{eqn:transformer_latency}, \ref{eqn:block_latency}), further making the program \textit{nonlinear}.

To solve this MINLP~\ref{eqn:program}, we leverage the Python numerical decomposition framework Pyomo~\cite{bynum2021pyomo} and MindtPy~\cite{bernal2018mixed}, and employ the Feasibility Pump (FP) method~\cite{bonami2009feasibility} to enhance efficiency. By jointly optimizing all variables, our approach achieves a globally optimal solution for $\bm{\omega}$ and $\bm{\kappa}$ in a single pass.

\noindent \textbf{Extract Pruned Structure} Once we solved the MINLP program~\ref{eqn:program}, we proceed to extract the pruned subnetwork $\hat{\Theta}$ based on the variables $\bm{\omega}$ and $\bm{\kappa}$ determined by the solver. If the block decision variable $\kappa_b$ is set to 0 for any block $b$, we completely remove that block from $\hat{\Theta}$, ignoring the solver’s output for the dimension variables $\bm{\omega}_b$ associated with that block. Conversely, if a block is active with $\kappa_b = 1$, we retain elements in each dimension $i$ of block $b$ according to the index of the ``hot" bit in $\omega_b^i$. Specifically, if the ``hot" bit is at index $j$, we retain the $j$ highest-ranking elements according to the Taylor channel score (for convolutions) or Hessian transformer score (for transformers) used when constructing the importance vector $\Vec{I}$.

After pruning, we finetune the model $\hat{\Theta}$ for $E$ epochs to restore any lost accuracy. The procedure is detailed step-by-step in Algorithm~\ref{algo:pseudocode}.

\begin{algorithm}[!t]
\small
    \caption{\method{} Framework}\label{euclid}
    \label{algo:1}
    \textbf{Input:} Pretrained weights $\Theta$, latency lookup table $T$, total finetuning epochs $E$, training dataset $\mathcal{D}$, latency budget $\Psi$
    \begin{algorithmic}[1]
    \State{Declare variables $\bm{\omega}$ and $\bm{\kappa}$}
        \State{\texttt{//Importance Objective Function}}
        \For{sample $(x, y)$ in $\mathcal{D}$}
        \State{Perform forward pass and backward pass with $\Theta$}
        \State{Compute CNNs or Transformer importance score} 
        \State{Construct importance vector $\Vec{I}$}
        \EndFor
        \State{Set up objective function $\mathcal{I}(\bm{\omega}, \bm{\kappa})$} (Eqn. \ref{eqn:importance}, \ref{eqn:block_importance})
        \State{\texttt{//Latency Constraint Function}}
        \State{Load prepared latency look-up tables $\mathbf{C}$}
        \State{Set up constraint function $\mathcal{C}'(\bm{\omega}, \bm{\kappa})$ (Eqn.\ref{eqn:conv_latency}, \ref{eqn:transformer_latency}, \ref{eqn:block_importance})}
        \State{\texttt{//Solve MINLP \& Extract $\hat{\Theta}$}}
        \State{Set up the MINLP (Eqn. \ref{eqn:program})} \State{Solve it with Pyomo and MindtPy}
        \State{Extract pruned $\hat{\Theta}$ from solver output $\bm{\omega}, \bm{\kappa}$}
        \State{Finetune the pruned model $\hat{\Theta}$ as usual for $E$ epochs}
    \end{algorithmic}
    \label{algo:pseudocode}
\end{algorithm}
\section{Experiments}
\begin{table}[th!]
    \centering
    \begin{minipage}{0.5\textwidth}
        \centering
        \resizebox{\textwidth}{!}
        {
        \begin{tabular}{lcccc}
            \toprule
            \rowcolor{lgray} 
            \textsc{Method} & \textsc{Top-1}$\uparrow$ & \textsc{Top-5}$\uparrow$  & \textsc{FLOPs}$\downarrow$ & \textsc{Speedup($\times$)}$\uparrow$ \\
            \rowcolor{lgray} 
             & ($\%$) & ($\%$)  & ($\times e^9$) &  \\
            \midrule
            \multicolumn{5}{c}{\textbf{ResNet50~\cite{he2016deep}}}\\
            \textsc{Dense} & $76.2$&$92.9$ & $4.1$&$1.00$\\
            \midrule
            \textsc{ResConv-Prn}\cite{xu2020layer} & $70.0$ & $90.0$ & $1.6$ & $-$ \\
            \textsc{DBP-0.5}\cite{wang2019dbp} & $72.4$ & $-$ & $-$ & $1.60^*$ \\
            \textsc{LayerPrune$_7$}\cite{elkerdawy2020filter} & $74.3$ & $-$ & $-$ & $1.79^*$ \\
            \textsc{MetaPrune}\cite{liu2019metapruning} & $73.4$ & $-$& $1.0$ & $2.34$ \\
            \textsc{AutoSlim}\cite{yu2019autoslim} & $74.0$ & $-$& $1.0$ & $2.35$ \\
            \textsc{GReg-2}\cite{wang2021neural} & $73.9$ & $-$& $1.3$ & $1.49$ \\
            \textsc{DCFF}\cite{lin2023training} & $73.8$ & $91.6$& $-$ & $2.23$ \\
            \textsc{HALP-$70\%$}\cite{shen2021halp} & $74.5$ &$91.8$ & $1.2$ & $2.55$ \\
            \textsc{SMCP-$70\%$}\cite{humble2022soft} & $74.6$ &$92.0$ & $1.0$ & $2.89$ \\
            \textsc{AHC-A}\cite{wang2024all} & $74.7$ &$92.1$ & $-$ & $2.24$ \\
            \textsc{DTP}\cite{li2023differentiable} & $74.3$ &$-$ & $-$ & $3.06$ \\
            \rowcolor{lgreen} \textbf{\textsc{Ours-70\%}} & $\mathbf{74.8}$ &$\mathbf{92.2}$ & $1.1$ & $\mathbf{3.06}$ \\
            \midrule
            \textsc{HALP-$85\%$}\cite{shen2021halp} & $68.1$ &$88.4$& $0.6$ & $3.90$ \\
            \rowcolor{lgreen} \textbf{\textsc{Ours-85\%}} & $\mathbf{70.0}$ &$\mathbf{89.3}$& $0.5$ & $\mathbf{5.21}$ \\
            \midrule
            \midrule
            \multicolumn{5}{c}{\textbf{ResNet50 - EagleEye~\cite{li2020eagleeye}}}\\
            \textsc{Dense}~\cite{li2020eagleeye} & $77.2$ &$93.7$& $4.1$&$1.00$\\
            \midrule
             \textsc{EagleEye-1G}\cite{li2020eagleeye} & $74.2$ &$91.8$ & $1.0$ & $2.38$ \\
             \textsc{HALP-70\%}\cite{shen2021halp} & $74.5$ &$91.9$ & $1.2$ & $2.65$ \\
              \textsc{SMCP-70\%}\cite{humble2022soft} & $75.1$ &$92.3$ & $1.1$ & $2.54$ \\
              \textsc{IEE-70\%}\cite{sun2025advancing} & $74.8$ &$92.2$ & $1.1$ & $2.57$ \\
                \rowcolor{lgreen}
              \textbf{\textsc{Ours-65\%}} & $\mathbf{75.2}$ &$\mathbf{92.5}$& $1.3$ & $\mathbf{2.72}$ \\
              \rowcolor{lgreen}
              \textbf{\textsc{Ours-70\%}} & $75.0$ &$92.2$& $1.2$ & $3.00$ \\
              \midrule
              \textsc{HALP-80\%}\cite{shen2021halp} & $71.2$ &$90.1$& $0.7$ & $3.62$ \\
              \textsc{SMCP-80\%}\cite{humble2022soft} & $72.7$ &--& --& $3.71$ \\
              \rowcolor{lgreen}
              \textbf{\textsc{Ours-80\%}} & $\mathbf{72.8}$ &$\mathbf{90.9}$& $0.7$ & $\mathbf{4.13}$ \\
              \midrule
            \textsc{HALP-85\%}\cite{shen2021halp} & $68.6$ &$88.5$& $0.6$ & $4.02$ \\
            \rowcolor{lgreen}
            \textbf{\textsc{Ours-85\%}} & $\mathbf{70.0}$ &$\mathbf{89.2}$& $0.5$ & $\mathbf{5.16}$ \\
            \bottomrule
        \end{tabular}
        }
        \label{tab:table1}
    \end{minipage}%
    \vspace{-5pt}
    \caption{\textbf{Classification on ImageNet}. \textit{Pruning CNN model ResNet-50}. Speed is measured on NVIDIA TITAN V with batch size $256$, and speedups are relative to dense FPS, with results grouped by similar speedups. $-X\%$ denote the pruning ratio. $^*$ denotes latency estimated from the reported. Our method shows superior accuracy-speed tradeoffs, especially at high pruning ratios. Results averaged over two runs.}
    \label{table:imageresnet50}
    \vspace{-12pt}
\end{table}

\begin{table}[th!]
    \centering
    \resizebox{0.9\linewidth}{!}
    {
        \begin{tabular}{lcc}
            \toprule
            \rowcolor{lgray} 
            \textsc{Method} & \textsc{Top-1}($\%$)$\uparrow$  & \textsc{Speedup}$\uparrow$ \\
            \midrule
            \textsc{DEIT-B}~\cite{touvron2021training} & $83.36$ & $1.00$ \\
            \textsc{SWIN-B}~\cite{liu2021swin} & $83.30$ & $0.95$ \\
            \textsc{NVIT-B}~\cite{yang2023global} & $83.29$ & $1.86$ \\
            \rowcolor{lgreen}
            \textbf{\textsc{Ours-$\mathbf{54\%}$}} & $\mathbf{83.47}$ & $\mathbf{1.87}$ \\
            \midrule
            \textsc{SWIN-S}~\cite{liu2021swin} & $83.00$ & $1.49$ \\
            \textsc{S$^2$ViTE-B-40}~\cite{chen2021chasing} & $82.92$ & $1.33$ \\
            \textsc{AutoFmer-B}~\cite{chen2021autoformer} & $82.90$ & $1.34$ \\
            \textsc{SAViT-H}~\cite{zheng2022savit} & $82.54$ & $1.55$ \\
            \textsc{NVIT-H}~\cite{yang2023global} & $82.95$ & $2.01$ \\
            \rowcolor{lgreen}
            \textbf{\textsc{Ours-$\mathbf{50\%}$}} & $\mathbf{83.29}$ & $\mathbf{2.03}$ \\
            \midrule
            \textsc{DEIT-S}~\cite{touvron2021training} & $81.20$ & $2.44$ \\
            \textsc{SWIN-T}~\cite{liu2021swin} & $81.30$ & $2.58$ \\
            \textsc{EViT-B}~\cite{liang2022not} & $82.10$ & $1.59$ \\
            \textsc{SAViT-S}~\cite{zheng2022savit} & $81.66$ & $2.05$ \\
            \textsc{CrossViT-S}~\cite{chen2021crossvit} & $81.00$ & $2.55$ \\
            \textsc{PVTv2-B2}~\cite{wang2022pvt} & $82.00$ & $2.21$ \\
            \textsc{NVIT-S}~\cite{yang2023global} & $82.19$ & $2.52$ \\
            \textsc{Isomorphic-S}~\cite{fang2025isomorphic} & $82.41$ & $2.56$ \\
            \rowcolor{lgreen}
            \textbf{\textsc{Ours-$\mathbf{39\%}$}} & $\mathbf{82.65}$ & $\mathbf{2.73}$ \\
            \midrule
            \textsc{DEIT-T}~\cite{touvron2021training} & $74.50$ & $5.18$ \\
            \textsc{AutoFmer-T}~\cite{chen2021autoformer} & $75.70$ & $4.59$ \\
            \textsc{NVIT-T}~\cite{yang2023global} & $76.21$ & $4.97$ \\
            \textsc{Isomorphic-T}~\cite{fang2025isomorphic} & $77.50$ & $5.20$ \\
            \rowcolor{lgreen}
            \textbf{\textsc{Ours-$\mathbf{19\%}$}} & $\mathbf{78.21}$ & $\mathbf{5.57}$ \\
            \bottomrule
        \end{tabular}
    }
    \vspace{-5pt}
    \caption{\textbf{Classification on ImageNet}. \textit{Pruning transformer model DEIT-Base.} Speed is measured on NVIDIA V100 with batch size of $256$. Speedups are relative to the FPS of the baseline DEIT-Base model, with results grouped by similar speedups. $-X\%$ denote the pruning ratio. Ours achieve much better accuracy-speed tradeoffs than prior art. Averaged results over two runs.
    }
    \label{table:transformer}
\vspace{-20pt}
\end{table}

To validate the proposed method, we perform extensive experiments across a comprehensive set of scenarios. We demonstrate our pruning results on both convolutional model(e.g. ResNet50~\cite{he2016deep}) and transformer architecture(e.g. DEIT-Base~\cite{touvron2021training}). Besides image classification on ImageNet~\cite{deng2009imagenet}, we also study 2D object detection with Pascal VOC~\cite{everingham2010pascal} and SSD~\cite{liu2016ssd}, and 3D object detection with Nuscenes~\cite{caesar2020nuscenes} and StreamPETR~\cite{wang2023exploring}. 


\noindent \textbf{Settings} We optimized ResNet50 and SSD for inference latency on an Nvidia TITAN V GPU (batch size 256). For DEIT-Base, we targeted latency reduction on an Nvidia V100 GPU (batch size 256) for fair comparison with methods like NViT. StreamPETR was evaluated on an NVIDIA RTX 3090 (batch size 1) to match its original setup. The appendix includes additional results on CPU. This demonstrates our method's adaptability across hardware platforms. Training was conducted on cluster with 8 Nvidia Tesla V100 GPUs using PyTorch V1.4.0, with an Intel Xeon E5-2698 CPU for solving the MINLP optimization(Eqn. ~\ref{eqn:program}). 

\subsection{CNN Pruning on ImageNet} Table~\ref{table:imageresnet50} compares our pruning results on ResNet50~\cite{he2016deep} and ImageNet~\cite{deng2009imagenet} with competitive baselines. We report Top-1 and Top-5 Accuracy to show accuracy recovery after fine-tuning, along with inference speedup relative to baseline ResNet50 for hardware performance gains. FLOPs are also included for completeness. 

Compared with previous methods like HALP~\cite{shen2021halp} and SMCP~\cite{humble2022soft}, we achieve a significantly improved accuracy-speed trade-off. For instance, SMCP reaches a Top-1 accuracy of $72.7$ with a speedup of $\times3.71$; our method slightly surpasses its Top-1 with an accuracy of $\mathbf{72.8}$ but with a considerably faster inference speed of $\mathbf{\times4.13}$. With larger pruning, HALP achieves a Top-1 accuracy of $68.6$ with an inference speedup of $\times4.02$, our method significantly outperforms it with a Top-1 accuracy of $\mathbf{70.0}$ and an impressive speedup $\times\mathbf{5.16}$. Notably, we can observe from Table~\ref{table:imageresnet50} that our method particularly excels when \textbf{\textit{targeting high speed}} with substantial pruning from pre-trained models, corroborating the effectiveness of improvements from our method. Our improvements could be observed more clearly in the Speedup v.s. Top-1 Pareto curve displayed in Figure~\ref{fig:teaser}. 

We also include direct comparison with specialized layer and block pruning methods~\cite{xu2020layer, wang2019dbp, elkerdawy2020filter}. Our results are significantly better, for instance, compared to LayerPrune~\cite{elkerdawy2020filter}, we achieve a higher Top-1 accuracy ($\mathbf{74.8}$ vs. $74.3$) and a substantial speedup ($\mathbf{\times3.06}$ vs. $\times1.79$).

\subsection{Transformer Pruning on ImageNet}
Table~\ref{table:transformer} shows our pruning results on DEIT-Base~\cite{touvron2021training}, compared against competitive transformer pruning methods and efficient transformer architectures on ImageNet~\cite{deng2009imagenet}. We assess performance using Top-1 accuracy and report efficiency as the speedup over the baseline DEIT-Base.

Our framework at every speedup level surpasses the established transformer pruning method NViT~\cite{yang2023global}, which we adopt for importance scoring, with especially large gains at higher speedups. For example, compared to NViT-T, our approach achieves higher accuracy ($\mathbf{78.21}$ vs. $76.21$) and significantly faster inference speed ($\mathbf{\times5.57}$ vs. $\times4.97$). Against the recent state-of-the-art method Isomorphic~\cite{fang2025isomorphic} (ECCV'24), we observe further improvement, with higher accuracy ($\mathbf{78.21}$ vs. $77.50$) and greater speedup ($\mathbf{\times5.57}$ vs. $\times5.20$). These results demonstrate our framework’s effectiveness in achieving leading accuracy-speed tradeoffs across both convolutional and transformer architectures.

\subsection{2D Object Detection Pruning on PascalVOC} 
\begin{figure}[t!]
    \centering
    \includegraphics[width=.8\linewidth]{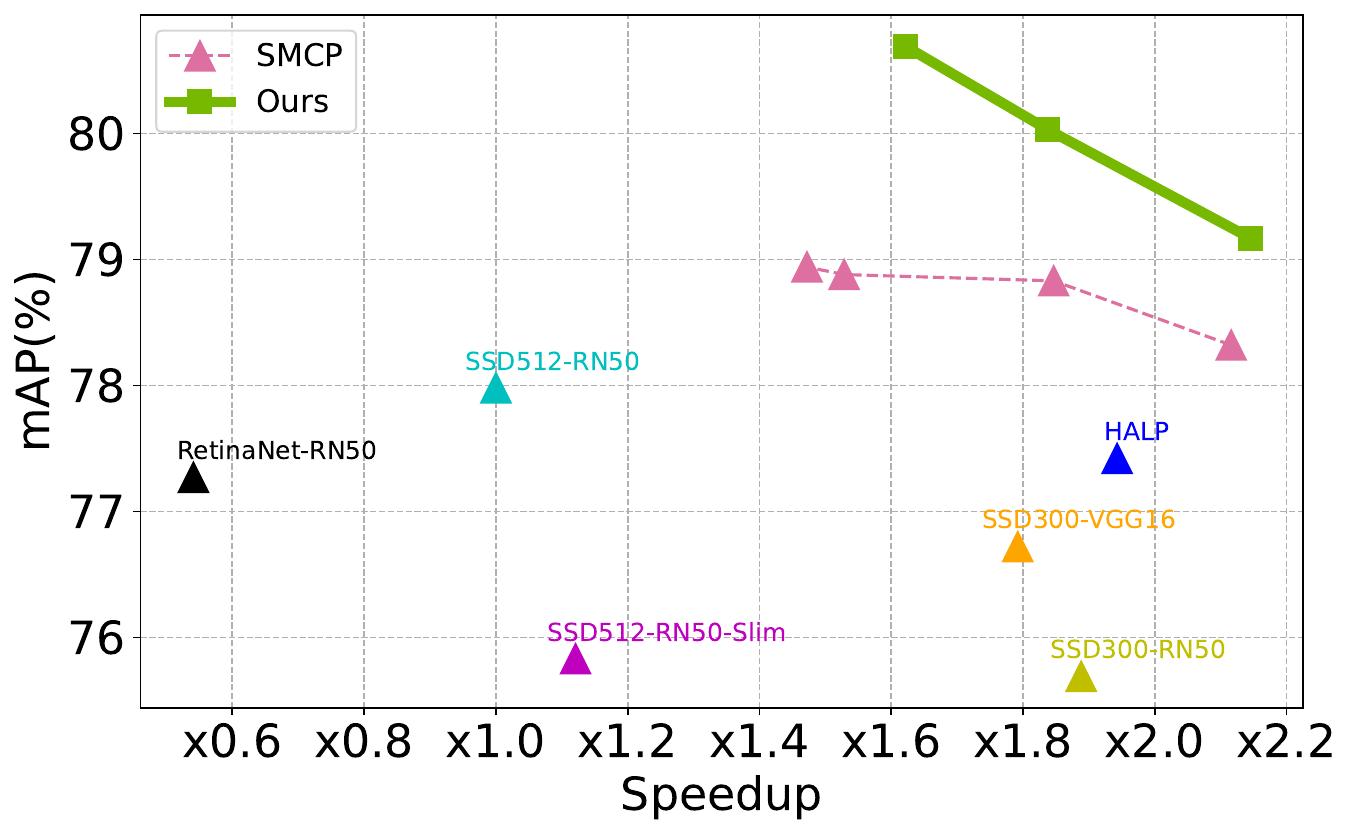}
    \vspace{-5pt}
    \caption{\textbf{2D Object Detection on Pascal VOC}. \textit{Pruning SSD512.} Speedup versus mAP are plotted(top-right is better). Speedup measured on NVIDIA TITANV, and is relative to the dense FPS.}
    \label{fig:pascal}
    \vspace{-20pt}
\end{figure}

To demonstrate our approach’s broad applicability, we ran experiments on 2D object detection with the Pascal VOC dataset~\cite{everingham2010pascal} pruning SSD512~\cite{liu2016ssd} with a ResNet50 backbone, with results shown in Figure~\ref{fig:pascal}. The Pareto frontier illustrates the trade-off between speedup (over baseline SSD512) and mean Average Precision (mAP).

Our results distinctly outshine existing methods in the field, marking a substantial advancement. In direct comparison to SMCP, our approach consistently achieves significantly higher mAP scores across various inference speed levels. For instance, we outperform SMCP with an mAP of $\mathbf{79.2}$ (compared to $78.3$) while also slightly increasing the speed to $\mathbf{\times2.14}$ (compared to $\times2.12$). Notably, our pruned model even surpasses the mAP of the pre-trained dense SSD512-RN50 by a margin($\mathbf{80.0}$ v.s. $78.0$)while substantially enhancing its speed($\times\mathbf{1.84}$ v.s. $\times1.00$).

\begin{table}[t!]
\centering
\resizebox{.85\linewidth}{!}
{
    \begin{tabular}{c|cc|c}
    \toprule
    \rowcolor{lgray}
    \textsc{Method} & \textsc{mAP$\uparrow$} & \textsc{NDS$\uparrow$} & \textsc{Speedup$\uparrow$} \\
    \midrule
    \textsc{BEVDet4D}\cite{huang2022bevdet4d}& $0.322$ & $0.457$ & $0.53$ \\
    \textsc{PETRv2}\cite{liu2023petrv2} & $0.349$ & $0.456$ & $0.60$ \\
    \textsc{Sparse4Dv2}\cite{lin2023sparse4d} & $0.439$ & $0.539$ & $0.64$ \\
    \textsc{StreamPETR}\cite{wang2023exploring} & $0.449$ & $0.546$ & $1.00$ \\
    \textsc{HALP-45\%}~\cite{shen2021halp} & $0.446$ & $0.547$ & $1.16$ \\
    \rowcolor{lgreen}
    \textbf{\textsc{Ours-45\%}} & $\mathbf{0.451}$ & $\mathbf{0.551}$ & $\mathbf{1.18}$ \\
    \midrule
    \textsc{HALP-50\%}~\cite{shen2021halp} & $0.439$ & $0.543$ & $1.21$ \\
    \rowcolor{lgreen}
    \textbf{\textsc{Ours-50\%}} & $\mathbf{0.441}$ & $\mathbf{0.544}$ & $\mathbf{1.23}$ \\
    \midrule
    \textsc{HALP-60\%}~\cite{shen2021halp} & $\mathbf{0.427}$ & $\mathbf{0.533}$ &$1.25$ \\
    \rowcolor{lgreen}
    \textbf{\textsc{Ours-60\%}} & $\mathbf{0.427}$ & $0.532$ & $\mathbf{1.28}$ \\
    \midrule
    \textsc{HALP-70\%}~\cite{shen2021halp} & $0.373$ & $0.489$ & $1.34$ \\
    \rowcolor{lgreen}
    \textbf{\textsc{Ours-70\%}} & $\mathbf{0.394}$ & $\mathbf{0.512}$ & $\mathbf{1.36}$ \\
    \bottomrule
    \end{tabular}
}
\vspace{-5pt}
\caption{\textbf{3D Object Detection on Nuscenes}. \textit{Pruning StreamPETR}. Speedup is measured on NVIDIA GeForce RTX 3090 with batch size of $1$, and relative to the dense FPS. Results with similar speedups are grouped. $-X\%$ denote the pruning ratio. Ours achieve much better accuracy-speed tradeoffs than HALP and even surpass performance of dense StreamPETR with higher speed.}
\label{table:nuscenes}
\vspace{-15pt}
\end{table}
\subsection{3D Object Detection Pruning on Nuscenes}
To demonstrate our approach’s applicability to 3D object detection, we tested on the challenging Nuscenes~\cite{caesar2020nuscenes} dataset using the state-of-the-art StreamPETR~\cite{wang2023exploring} model. Table~\ref{table:nuscenes} presents detailed results and comparisons with competitive baselines. Our evaluation includes standard metrics like mean Average Precision (mAP) and Normalized Detection Score (NDS), along with speedup relative to the base StreamPETR model to emphasize efficiency gains.

Significantly, when compared to the dense pre-trained StreamPETR model, our technique achieved a substantial acceleration of approximately $\mathbf{18\%}$, resulting in an impressive $\times\mathbf{1.18}$ speedup as opposed to the baseline. Importantly, this speed boost was achieved without sacrificing performance: our pruned model attained superior mAP ($\mathbf{0.451}$ vs. $0.449$) and NDS ($\mathbf{0.551}$ vs. $0.546$). In comparison to the previous pruning method HALP~\cite{shen2021halp}, our approach exhibited remarkable improvements in accuracy-latency trade-offs especially at high pruning ratios. For instance, HALP managed to produce a pruned StreamPETR model with an mAP of $0.373$, an NDS of $0.489$, and an inference speedup of $\times1.34$. In contrast, our approach surpassed these results, achieving an mAP of $\mathbf{0.394}$, an NDS of $\mathbf{0.512}$, and an inference speedup of $\mathbf{\times1.36}$.

\subsection{Ablation Study} 
As discussed in Sec.~\ref{sec:intro}, our multidimensional pruning method introduces two key advancements: \textbf{MGP}(\textbf{\textit{Multi-Granularity Pruning}}) for block-level pruning, and \textbf{MDLM}(\textbf{\textit{Multi-Dimensional Latency Modeling}}) for accurately capturing latency across all prunable dimensions.

\begin{figure}[t!]
    \centering
    \includegraphics[width=.85\linewidth]{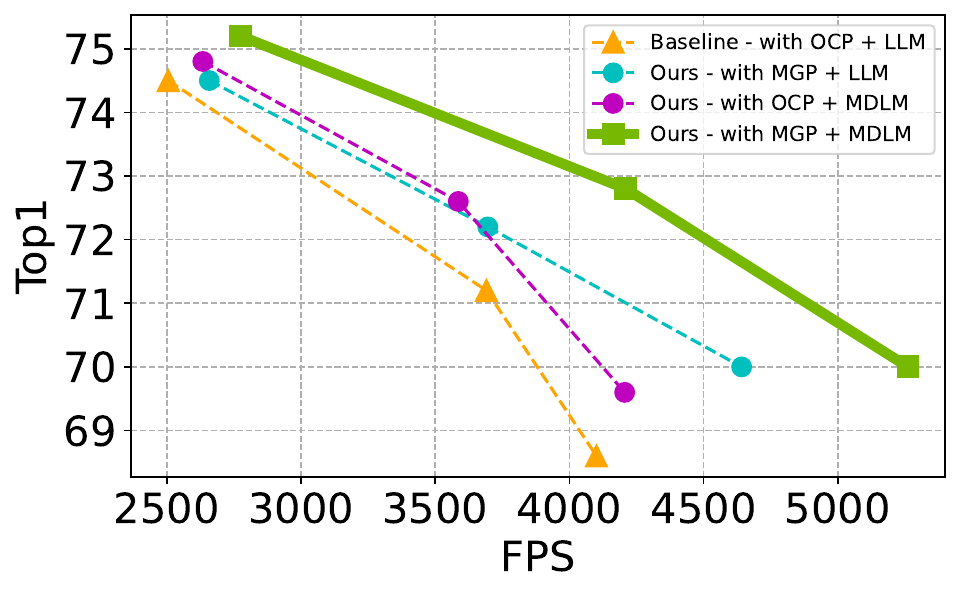}
    \vspace{-10pt}
    \caption{\textbf{Ablation study results on ImageNet with ResNet50}. We show results of each improvement acting individually. Top-right is better. \textbf{MGP: Multi-Granularity Pruning (Ours)}; \textbf{MDLM: Multi-Dimensional Latency Modeling (Ours)}; OCP: Only Channel Pruning; LLM: Linear Latency Modeling}
    \label{fig:ablation}
    \vspace{-18pt}
\end{figure}

To assess the individual impact of MGP and MDLM, we start with a baseline pruning algorithm that lacks both components. Here, the baseline uses \textbf{OCP}(\textbf{\textit{Only Channel Pruning}}) as an alternative to our MGP, and replaces our MDLM with \textbf{LLM} (\textbf{\textit{Linear Latency Modeling}}), which was proposed in prior works \cite{shen2021halp, humble2022soft, shen2023hardware} and only linearly considers output channel counts in CNNs. We then add MGP and MDLM separately to evaluate their contributions and compare against prior methods. The baseline performance, labeled ``\textit{Baseline} - OCP+LLM,'' is shown in Fig.~\ref{fig:ablation}, while our proposed \textbf{MDP} paradigm with both components is labeled as ``\textit{Ours} - MGP+MDLM''.

\noindent\textbf{MGP+LLM:} This variant incorporates block-level pruning but does not accurately capture latency variations across all dimensions, including input and output channels. Results, labeled ``\textit{Ours - MGP+LLM}'' in Fig.~\ref{fig:ablation}, show a significantly improved accuracy-latency tradeoff over the baseline, particularly at high pruning ratios and latency reductions (rightmost points). This underscores the effectiveness of Multi-Granularity Pruning alone. However, compared to \textit{MGP+MDLM}, it is outperformed at all levels, highlighting the value of Multi-Dimensional Latency Modeling.

\noindent\textbf{OCP+MDLM:} This variant models latency accurately with MDLM but lacks the ability to remove entire blocks. Results, labeled ``\textit{Ours - OCP+MDLM}'' in Fig.~\ref{fig:ablation}, reveal a clear improvement in the accuracy-latency tradeoff compared to the baseline, confirming the effectiveness of Multi-Dimensional Latency Modeling. Nevertheless, it performs worse than \textit{MGP+MDLM}, emphasizing the importance of Multi-Granularity Pruning.

\vspace{-3pt}
\section{Conclusion}
In this paper, we present \method{}, a novel paradigm that unifies combined-granularity pruning within a single optimization process and introduces a precise latency modeling technique to account for simultaneous variations across all prunable dimensions. To incorporate these strategies, we reformulate pruning as a Mixed-Integer Nonlinear Program (MINLP), enabling efficient identification of the optimal pruned structure within a latency budget across all dimensions in a single pass. Our results show substantial improvements over prior methods, particularly in scenarios with large pruning. Additionally, we provide an ablation study to analyze the impact of each component of our framework.
\maketitlesupplementary
\setcounter{page}{1}

\section{Latency Modeling: Ours v.s. Prior Arts}
In the main paper, we highlight that prior approaches~\cite{shen2021halp, shen2023hardware, humble2022soft} rely on imprecise latency estimation, leading to suboptimal accuracy-latency trade-offs. These methods use a simplistic latency model that assumes a linear relationship between latency and the number of output channels. However, this approach is limited as it cannot simultaneously account for multiple prunable dimensions and is restricted to CNNs.

\begin{figure}[t!]
\begin{center}
   \includegraphics[width=\linewidth]{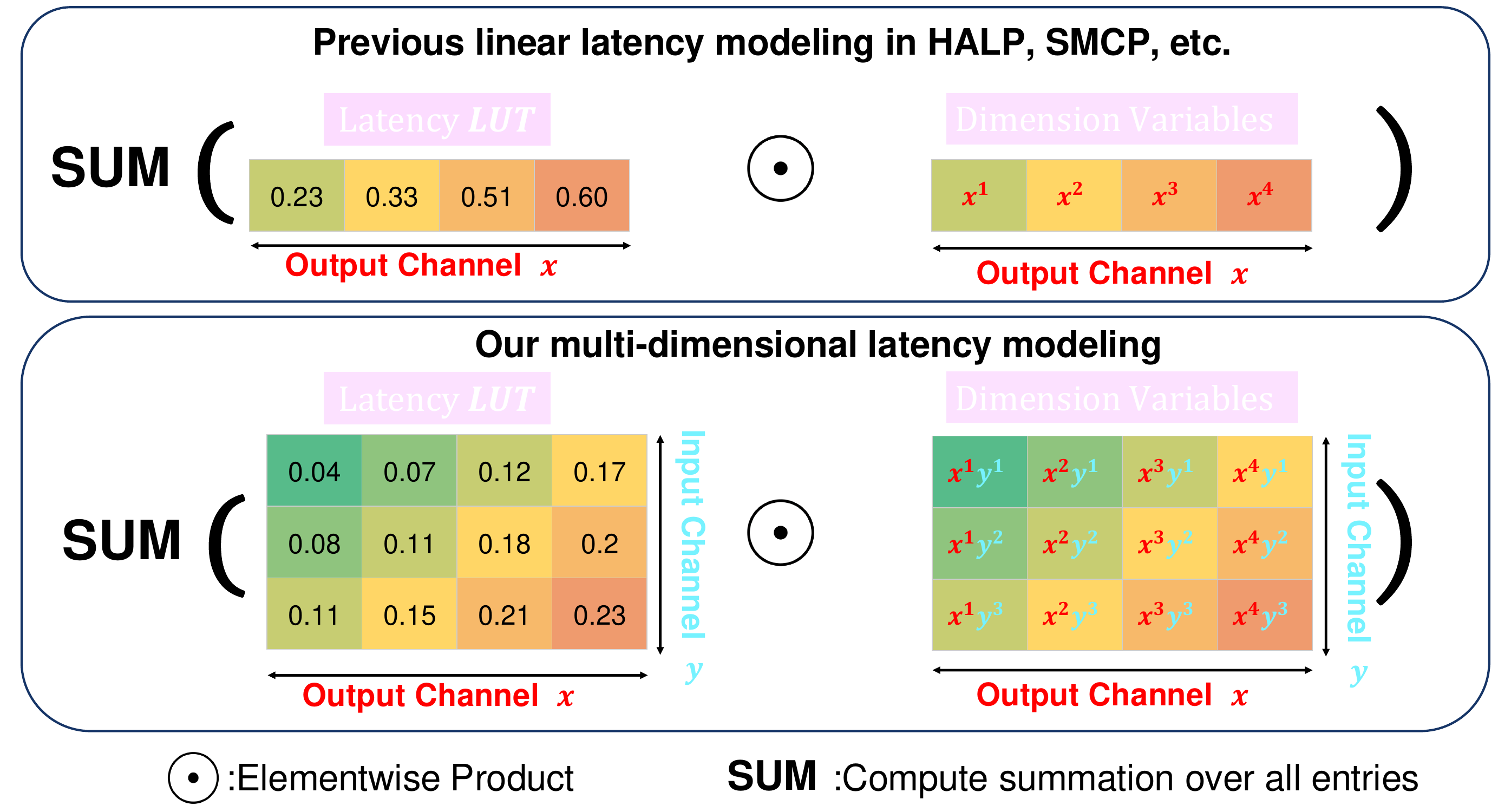}
\end{center}
\vspace{-5pt}
\caption{\textbf{Comparison in latency modeling between ours and prior arts~\cite{shen2021halp, humble2022soft}}. Example with CNNs.}
\label{fig:latencymodel}
\end{figure}

When pruning transformers, it is essential to model simultaneous variations across multiple dimensions, such as embedding size, query/key dimensions, value dimensions, number of heads, and MLP size, which cannot be captured by a linear model. Even for CNNs, such linear modeling lacks precision. To illustrate, we provide a visualization in Fig.~\ref{fig:latencymodel}, demonstrating the limitations of prior latency modeling. Specifically, these models only account for changes in output channels while neglecting concurrent variations in input channels caused by pruning in preceding layers.

\section{Solving MINLPs}
In order to solve our MINLP program, we leverage the method called OA~\cite{duran1986outer,fletcher1994solving} which decomposes the problem into solving an alternating finite sequence of NLP subproblems and relaxed versions of MILP master program. We also leverage a method called Feasibility Pump~\cite{bonami2009feasibility} to   to expedite the process of finding feasible solutions within constraints. 

The entire program could be efficiently solved on common CPUs for modern network sizes. For instance, when applied to a model like ResNet50~\cite{he2016deep}, the entire optimization problem can be solved in approximately 5 seconds on an Intel Xeon E5-2698 CPU. Moreover, in Table \ref{tab:efficiency}, we show the overhead of solving MINLPs for ResNet50 and DEIT-Base relative to their training times. As shown, solving the MINLP program is efficient, for example taking only $\mathbf{0.1\%}$ of DEIT-Base’s total training time. Theoretically, as established in ~\cite{lee2011mixed}, the problem remains tractable and scalable as long as the objective and constraint functions are convex separable or gated with a switch variable. Here, block decision variables act as gates, simplifying the solving.

\section{Efficiency of LUTs Preparation}
In our framework, we leverage latency look-up tables (LUTs) to model the latency impacts from different pruning decisions. In LUTs, the inference latency for \textit{all possible} local subnet structures are recorded. For CNN layers for example, this includes input and output channel counts from $0$ up to their original values. While this may seem computationally expensive, the process is efficient in practice. Firstly, LUTs are generated \textit{only once} to measure target hardware latency and can be reused for future pruning runs, with minimal overhead even without parallelization (Table~\ref{tab:efficiency}). Moreover, our \textit{model-specific decomposition} as shown in Eqn.\ref{eqn:conv_latency},\ref{eqn:transformer_latency} significantly reduces the total number of LUT entries, and we further adopt channel grouping~\cite{shen2021halp, humble2022soft, sun2025advancing}
to cluster channels of similar importance into single elements, minimizing overhead. Additionally, LUTs can be reused across related architectures (e.g., ResNet101 shares many LUT entries with ResNet50).

\section{Adaptation to CPU}
To demonstrate adaptability, we evaluated our method on the \textbf{CPU} platform using an Intel Xeon E5 processor. The results, summarized in Table \ref{table:cpu}, indicate substantial performance gains over prior work. Notably, compared to HALP~\cite{shen2021halp}, our method achieves more than double the FPS (\textbf{118.2} vs. 45.9) while also attaining a higher Top-1 accuracy (\textbf{75.2} vs. 74.5). This improvement is even more pronounced than the speedup observed on GPU. We attribute this to the reduced number of blocks and the smaller overall network depth, which make the network more CPU-friendly. These results highlight the effectiveness of MDP in generalizing across both CPU and GPU platforms.

\begin{table}[t!]
    \centering
    \resizebox{.85\linewidth}{!}{
    \begin{tabular}{cc|c}
        \toprule
        \rowcolor{lgray}
         \textsc{Train (mins)} & \textsc{Solve MINLP (mins)} & \textsc{LUT Prep. (mins)} \\
         \midrule
         \multicolumn{3}{c}{\textsc{\textbf{ResNet50} trained for 90  epochs}}\\
         $667 (\times 1)$ &$0.09(\times 0.01\%)$ & $152 (\times 23\%)$\\
         \midrule
         \multicolumn{3}{c}{\textsc{\textbf{DEIT-Base} trained for 300 epochs}}\\
         $7614 (\times 1)$ &$7.53(\times 0.1\%)$ & $320 (\times 4.2\%)$\\
         \bottomrule
    \end{tabular}
    }
    \caption{Overhead of solving MINLP and preparing the Look-up Table(LUT) for ResNet50 and DEIT-B. LUT only needs to be generated once. CPU in use is Intel Xeon E5-2698.}
    \label{tab:efficiency}
\end{table}

\begin{table}[ht!]
    \centering
    \resizebox{.8\linewidth}{!}
    {
        \begin{tabular}{c|cc}
            \toprule
            \rowcolor{lgray}
            \textsc{Method} & \textsc{Top-1}$\uparrow$ & \textsc{FPS}$\uparrow$\\
            \textsc{Autoslim}~\cite{yu2019autoslim} & $74.0$  & $33.3$\\
            \textsc{EagleEye}~\cite{li2020eagleeye} & $74.2$  & $31.3$\\
            \textsc{MetaPrune}~\cite{liu2019metapruning} & $73.4$  & $33.3$\\
            \textsc{HALP}-$70\%$~\cite{shen2021halp} & $74.5$  & $45.9$\\
            \rowcolor{lgreen}
            \textbf{Ours} & $\mathbf{75.2}$ & $\mathbf{118.2}$ \\
            \bottomrule
        \end{tabular}
    }
    \captionof{table}{\textbf{Generalization of MDP on CPU Platform.} FPS measured on Intel CPU Xeon E5. Ours attains significant improvements from prior arts, specifically in speedups.}
    \label{table:cpu}
\end{table}

\begin{table*}[t!]
\vspace{-8pt}
    \centering
    \resizebox{.99\linewidth}{!}{
    \begin{tabular}{cc|ccc}
        \toprule
        \rowcolor{lgray}
        Model & Dataset & Epochs & Optimizer, Momentum, WeightDecay & Learning Rate\\
        \hline
        ResNet50 & ImageNet & $90$  & SGD, $0.875$, $3e-5$ &Init=$1.024$, LinearDecay\\
        DEIT-Base & ImageNet & $300$  & AdamW, $0.9$, $0.05$ &Init=$2e-4$, CosineAnneal\\
        SSD512 & PascalVOC & $800$ & SGD, $0.9$, $2e-3$ &Init=$8e-3$, StepDecay \\
        StreamPetr & NuScenes & $60$  & AdamW, $0.9$, $0.01$ & Init=$6e-4$, CosineAnneal \\
        \bottomrule
    \end{tabular}
    }
    \vspace{-10pt}
    \caption{Training Detail.}
    \label{tab:detail}
    \vspace{-12pt}
\end{table*}

\begin{figure}
    \centering
    \includegraphics[width=\linewidth]{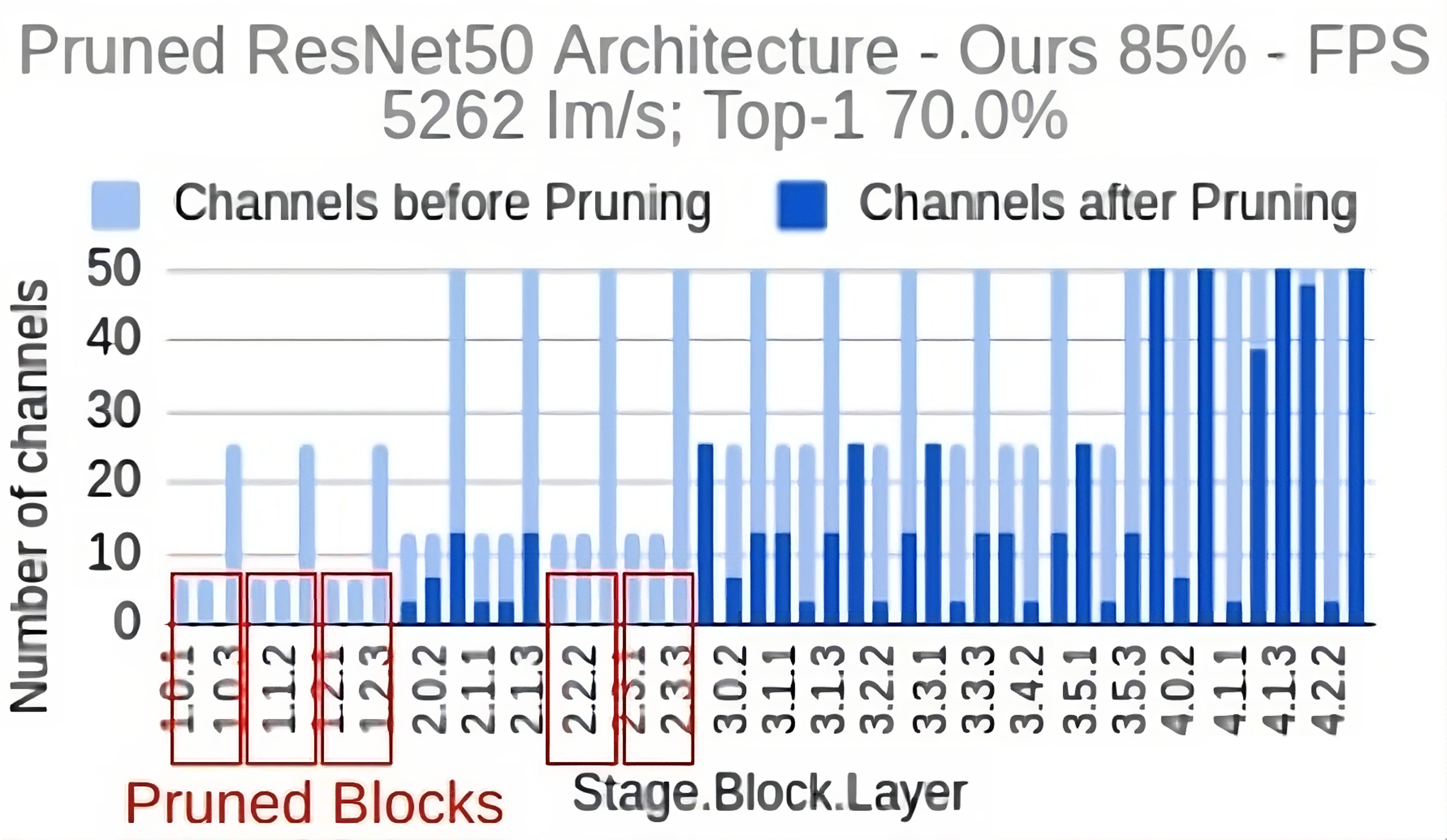}
    \caption{Pruned architecture of ResNet50 on ImageNet.}
    \label{fig:pruned_arch}
\end{figure}

\begin{figure}[t!]
\begin{center}
   \includegraphics[width=\linewidth]{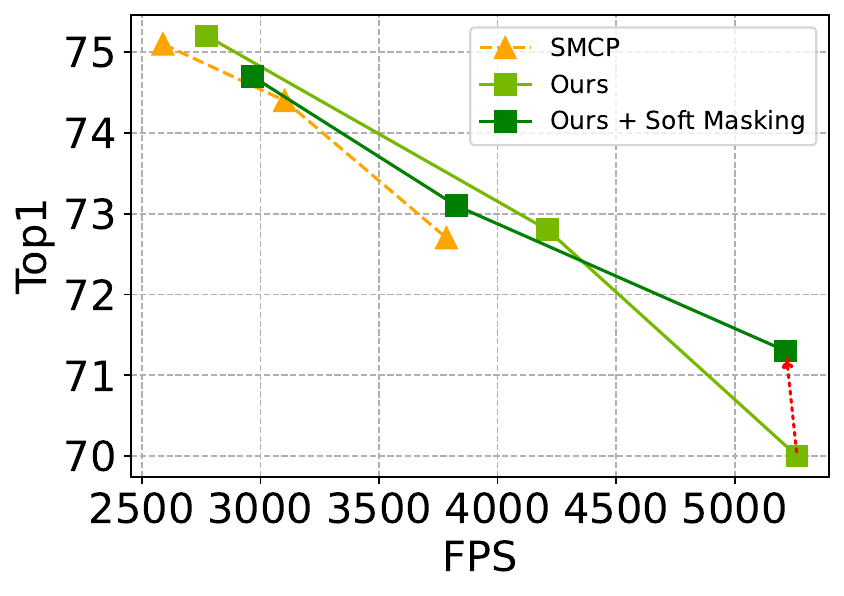}
\end{center}
\vspace{-5pt}
\caption{\textbf{Results of ours with soft masking on ImageNet with ResNet50}. We investigate the effectiveness of soft masking techniques in our method and observe improvement in Top1 at a high FPS level. Top-right is better.}
\label{fig:softmaskg}
\end{figure}

\begin{table}[t!]
    \centering
    \begin{minipage}{0.4\linewidth}
        \centering
        \resizebox{\linewidth}{!}{
        \begin{tabular}{c|cc}
            \toprule
            \rowcolor{lgray}
             \textsc{Model} & \textsc{Top-1} & \textsc{Speedup} \\
             \midrule
             MobileNet-V1 &$72.6$ & $\times 1.00$\\
             MetaPrune[\textcolor{cvprblue}{49}] &$66.1$ & $\times 2.06$\\
             HALP-42\%[\textcolor{cvprblue}{57}] &$68.3$ & $\times 2.32$\\
             SMCP-40\%[\textcolor{cvprblue}{34}] &$68.3$ & $\times 2.39$\\
             \rowcolor{lgreen}
             Ours &$\mathbf{68.5}$ & $\times \mathbf{2.41}$\\
             \midrule
             \midrule
             MobileNet-V2-1.4 &$75.3$ & $\times 1.00$\\
             MBV2-1.4-DS-E &$72.2$ & $\times 2.50$\\
             UPDP-P11~\cite{liu2024updp} & $72.5$ & $\times 2.50$\\
             \rowcolor{lgreen}
             Ours & $\mathbf{72.6}$ & $\times \mathbf{2.63}$\\
             \bottomrule
        \end{tabular}
        }
        \vspace{-8pt}
        \caption{Results on pruning efficient CNNs.}
    \label{tab:mobilenet}
        \label{tab:ampere}
    \end{minipage}%
    \hfill
    \begin{minipage}{0.57\linewidth}
        \centering
        \includegraphics[width=\linewidth]{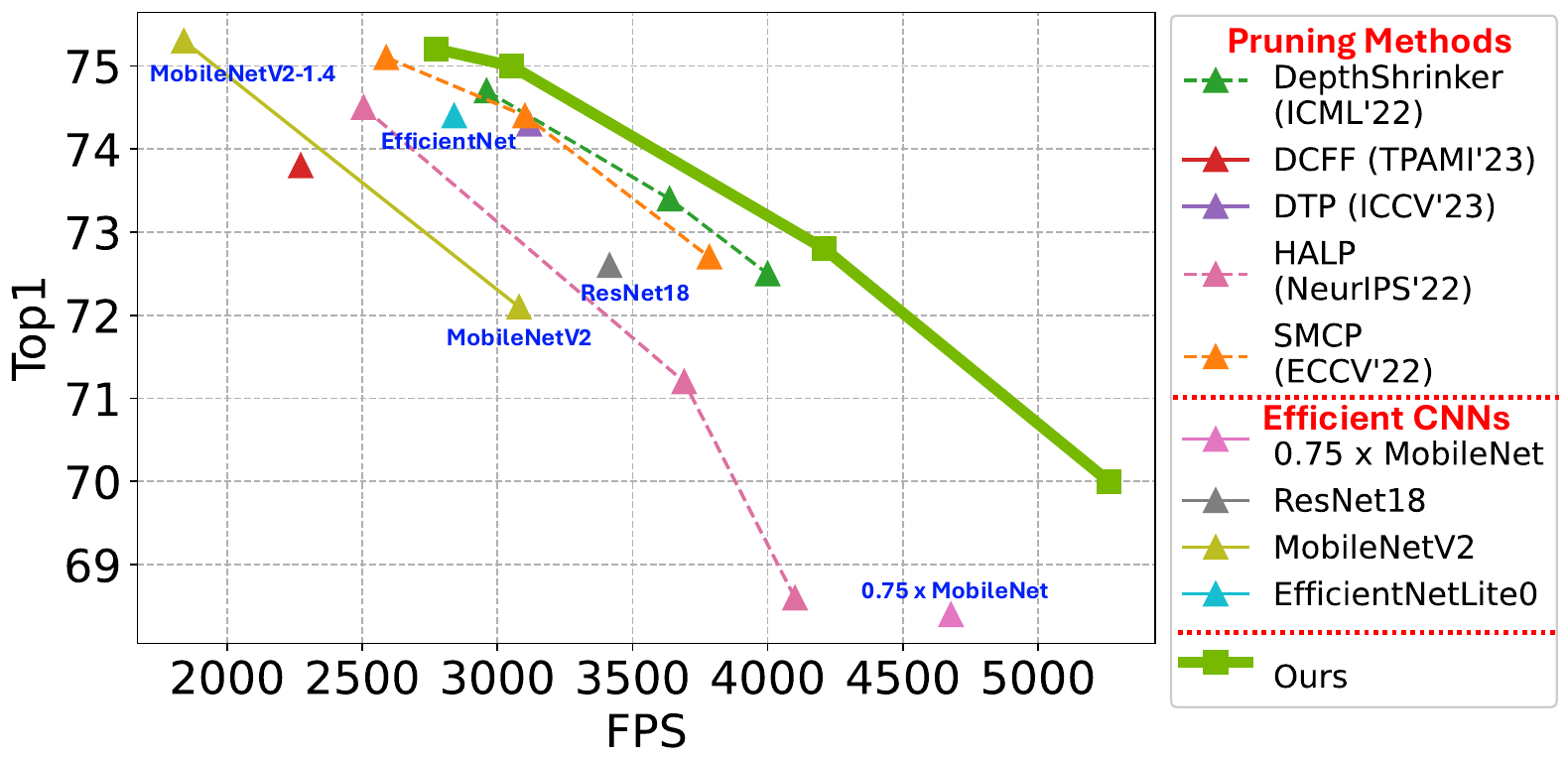}
        \vspace{-20pt}
        \captionof{figure}{Top-right is better. Recent pruning works surpass efficient CNNs in speed-accuracy tradeoffs on ImageNet, with ours achieving the best.}
        \label{fig:comparison}
    \end{minipage}
    \vspace{-10pt}
\end{table}

\section{Pruning Efficient CNNs}
We include results for pruning efficient CNNs like MobileNet-V1 and MobileNet-V2 in Table~\ref{tab:mobilenet}. Our method consistently outperforms prior approaches. For example, on MobileNet-V2, compared to UPDP~\cite{liu2024updp}, we achieve slightly higher Top-1 ($\mathbf{72.6}$ vs. $72.5$) and improved speed ($\mathbf{2.63}$ vs. $2.50$). 

Furthermore, our approach focuses on aggressively pruning moderate or large models to derive efficient ones. This strategy offers superior accuracy-speed tradeoffs compared to directly training efficient models from scratch, as shown in Fig.~\ref{fig:comparison}.

\section{Pruned Structure Analysis}
To provide insights into our pruning algorithm, we present the pruned structure of ResNet50 on ImageNet in Fig.~\ref{fig:pruned_arch}, targeting an 85\% latency reduction. The figure shows that pruning is predominantly concentrated in the shallower layers, contrary to the common expectation of deeper layers collapsing due to smaller gradients. This indicates that when latency is strict constraint, the pruning pattern is influenced not only by importance ranking but also by latency considerations. Since earlier layers process larger feature maps and are generally more latency-intensive, they are pruned more aggressively than later stages.

\section{Training Detail}
For reproducibility, we provide detailed hyperparameters and fine-tuning optimization settings in Table~\ref{tab:detail}, adhering to the baseline configurations. Specifically, when fine-tuning the pruned DEIT-Base model on ImageNet, we incorporate the distillation loss from the convolutional RegNetY160 model, as described in the original paper.

\section{Integration with Soft Masking}
Recent advances in pruning~\cite{humble2022soft, zhou2021learning, he2018soft, kusupati2020soft, kim2021dynamic} have increasingly adopted soft masking techniques to retain the learning capacity of pruned models by not directly removing the pruned weights. Notably, SMCP~\cite{humble2022soft} integrates this method into the HALP hardware-aware pruning framework, resulting in an enhanced accuracy-latency tradeoff for pruned models. Here, we explore the potential of soft masking to enhance our model's performance. 

We conduct this study on ImageNet with ResNet50 and depict the Pareto frontier of FPS versus Top-1 in Figure~\ref{fig:softmaskg}. For clarity, we also include the performance of SMCP~\cite{humble2022soft} and ours. The results reveal that soft masking offers limited advantages at lower FPS levels with modest pruning ratios and latency reduction. Nonetheless, targeting higher FPS levels leads to notable improvements in Top-1 accuracy. This outcome may be attributed to the Taylor channel importance score we employed~\cite{molchanov2019importance}, which gauges parameter significance based on its impact on loss. Though it maintains precision with minor parameter deletions, its reliability may diminish when a larger number of parameters are pruned concurrently. The iterative reassessment inherent to the soft masking technique may counteract this shortcoming.
%
%
{
    \small
    \bibliographystyle{ieeenat_fullname}
    \bibliography{main}
}
\end{document}